\documentclass{article}
\usepackage{spconf,amsmath,graphicx}
\usepackage{amssymb}
\usepackage{multirow}
\usepackage{balance}
\usepackage{bbm}
\usepackage{fancyhdr}

\title{Pair-level Supervised Contrastive Learning for Natural Language Inference}
\name{Shu'ang Li$^1$, Xuming Hu$^1$, Li Lin$^1$, Lijie Wen$^2$\sthanks{Corresbonding Author} }
\address{School of Software, Tsinghua University\\$^1$\{lisa18, hxm19, lin-l16\}@mails.tsinghua.edu.cn, $^2$wenlj@tsinghua.edu.cn}
\begin{document}
\maketitle
\thispagestyle{fancy}
\fancyhead{}
\lhead{}
\lfoot{
\fbox{%
\parbox{\textwidth}{%
    \ninept
    \copyright~2022 IEEE.  Personal use of this material is permitted. Permission from IEEE must be obtained for all other uses, in any current or future media, including reprinting/republishing this material for advertising or promotional purposes, creating new collective works, for resale or redistribution to servers or lists, or reuse of any copyrighted component of this work in other works.
}%
}
}
\cfoot{}
\rfoot{}
\begin{abstract}
Natural language inference (NLI) is an increasingly important task for natural language understanding, which requires one to infer the relationship between the sentence pair (\textbf{premise} and \textbf{hypothesis}).
Many recent works have used contrastive learning by incorporating the relationship of the sentence pair from NLI datasets to learn sentence representation. 
However, these methods only focus on comparisons with sentence-level representations.
In this paper, we propose a \textbf{Pair}-level \textbf{S}upervised \textbf{C}ontrastive \textbf{L}earning approach (PairSCL). We adopt a cross attention module to learn the joint representations of the sentence pairs. A contrastive learning objective is designed to distinguish the varied classes of sentence pairs by pulling those in one class together and pushing apart the pairs in other classes. We evaluate PairSCL on two public datasets of NLI where the accuracy of PairSCL outperforms other methods by 2.1\% on average. Furthermore, our method outperforms the previous state-of-the-art method on seven transfer tasks of text classification.
\end{abstract}
\section{Introduction}
\label{sec:intro}
Natural Language Inference (NLI) is a fundamental problem in the research ﬁeld of natural language understanding~\cite{maccartney-manning-2008-modeling,shen2021towards}, which could help tasks like questions answering, reading comprehension, summarization and relation extraction~\cite{dagan2013recognizing, hu2020selfore, hu2021gradient, hu-etal-2021-semi-supervised}. In NLI settings, the model is presented with a pair of sentences, namely \textbf{premise} and \textbf{hypothesis} and is asked to reason the relationship between them from a set of relationships, including \texttt{entailment}, \texttt{contradiction} and \texttt{neutral}. 
In the last several years, large annotated datasets were made available, e.g., the SNLI~\cite{bowman-etal-2015-large} and MultiNLI datasets~\cite{williams2018broad}, which made it feasible to train rather complicated neural network-based models~\cite{chen-etal-2017-enhanced,gong2018natural}. 


However, these methods only use the feature of the sentence pair itself to predict the class, without considering the comparison between the sentence pairs in different classes. Many recent works explored using contrastive learning to tackle this problem. Contrastive learning is a popular technique in computer vision area \cite{he2020momentum, chen2020simple, khosla2020supervised} and the core idea is to learn a function that maps positive pairs closer together in the embedding space, while pushing apart negative pairs. A contrastive objective is used by \cite{gao2021simcse} to ﬁne-tune pre-trained language models to obtain sentence embeddings with the relationship of sentences in NLI, and achieved state-of-the-art performance in sentence similarity tasks. However, this approach can't distinguish well between the representation of sentence pairs in different classes. For example, two sentence pairs are in the same class of \texttt{entailment} from NLI dataset ($P_1$: Two men on bicycles competing in a race. $H_1$: People are riding bikes. $P_2$: Two dogs are running. $H_2$: There are animals outdoors). They simply consider $H_1$ as the positive set and $H_2$ as the negative set for $P_1$ without taking into account that these two pairs are in the same class. 

Given this scenario, we propose a pair-level supervised contrastive learning approach. The pair-level representation is obtained by cross attention module which can capture the relevance and well characterize the relationship between the sentence pair. Therefore, the pair-level representation can perceive the class information of sentence pairs. Then we use the pair-level representations for contrastive learning by capturing the similarity between pairs in one class and contrasting them with pairs in other classes. The model is trained with a combined objective of a supervised contrastive learning term and a cross-entropy term. We evaluate PairSCL on two public datasets of NLI where the accuracy of PairSCL outperforms other methods by 2.1\% on average. Furthermore, our method outperforms the previous state-of-the-art method on seven transfer tasks of text classification.

\section{Approach}
\label{sec:model}

\begin{figure}[ht]
  \centering
  \includegraphics[width=0.77\linewidth]{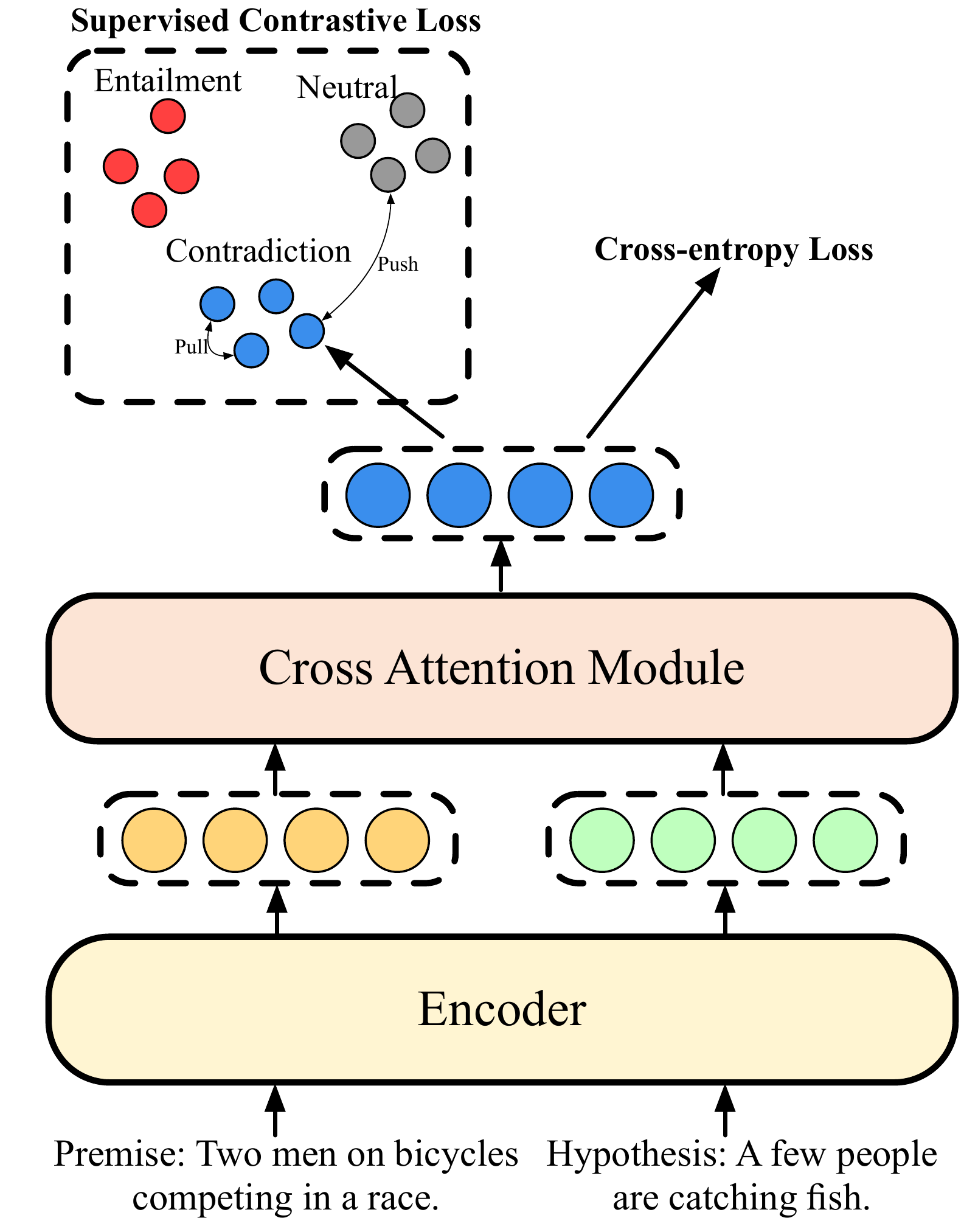} 
  \caption{\label{model} The framework of PairSCL.} 
\end{figure}
\vspace{-0.1in}

In this section, we describe our approach PairSCL. Figure \ref{model} shows a high-level general view of PairSCL. PairSCL comprises the following three major components: an encoder that computes sentence representations for input text, a cross attention module to capture the relationship between the sentence pair and a joint-training layer including a cross-entropy term and supervised contrastive learning term.

\subsection{Text Encoder}
\label{sec:encoder}
Each instance in a NLI dataset consists of two sentences and a label indicating the relation between them. Formally, we denote \textbf{premise} as $X^{(p)}=\{x_1^{(p)},x_2^{(p)},\cdots,x_m^{(p)}\}$ and \textbf{hypothesis} as $X^{(h)}=\{x_1^{(h)},x_2^{(h)},\cdots,x_n^{(h)}\}$, where $m$ and $n$ are length of the sentences respectively. The instance in the batch $\mathcal{I}$ is denoted as $(X^{(p)}, X^{(h)}, y)_{i \in \mathcal{I}}$, where $i = \{1, \dots, K\} $ is the indices of the samples and $K$ is the batch-size. The encoder (e.g., BERT, RoBERTa) takes $X^{(p)}, X^{(h)}$ as inputs and computes the semantic representations, denoted as $\mathbf{S}^{(p)} = \{\mathbf{s}_i^{(p)}|\mathbf{s}_i^{(p)} \in \mathbb{R}^{k}, i = 1,2,\cdots, m\}$ and $\mathbf{S}^{(h)} = \{\mathbf{s}_j^{(h)}|\mathbf{s}_j^{(h)} \in \mathbb{R}^{k}, j = 1,2,\cdots, n\}$, where $k$ is the dimension of the encoder’s hidden state.

\subsection{Cross Attention Module}

Different from single sentence classification, we need a proper interaction module to better clarify the sentences pair's relationship for NLI task. In practice, we need to compute token-level weights between words in \textbf{premise} and \textbf{hypothesis}. Therefore, we introduce the cross attention module to calculate the co-attention matrix $\mathbf{C} \in \mathbb{R}^{m \times n}$ of the token level. Each element $\mathbf{C}_{i,j} \in \mathbb{R}$ indicates the relevance between the i-th word of \textbf{premise} and the j-th word of \textbf{hypothesis}:
\begin{equation}
  \mathbf{C}_{i,j} = \mathbf{P}^Ttanh(\mathbf{W}(\mathbf{s}^{(p)}_i \odot \mathbf{s}^{(h)}_j)),
\end{equation}
where $\mathbf{W} \in \mathbb{R}^{d \times k}$, $\mathbf{P} \in \mathbb{R}^{d}$, and $\odot $ denotes the element-wise production operation. Then the attentive matrix could be formalized as:

\begin{align}
  \mathbf{c}_i^{(p)} = softmax(\mathbf{C}_{i,:}),\quad &\mathbf{c}_j^{(h)} = softmax(\mathbf{C}_{:,j}),\\
  \mathbf{s}^{(p)'}_i = \mathbf{S}^{(h)} \cdot \mathbf{c}_i^{(p)},\quad &\mathbf{s}^{(h)'}_i = \mathbf{S}^{(p)} \cdot \mathbf{c}_j^{(h)},
\end{align}

We further enhance the collected local semantic information:
\begin{gather}
  \mathbf{s}^{(p)''}_i = [\mathbf{s}^{(p)}_i; \mathbf{s}^{(p)'}_i; \mathbf{s}^{(p)}_i - \mathbf{s}^{(p)'}_i; \mathbf{s}^{(p)}_i \odot \mathbf{s}^{(p)'}_i],\\
  \tilde{\mathbf{s}}^{(p)}_i = ReLU(\mathbf{W}^{(p)}_i\mathbf{s}^{(p)''}_i + \mathbf{b}^{(p)}_i),
\end{gather}
where $[\cdot ;\cdot ;\cdot ;\cdot ]$ refers to the concatenation operation. $\mathbf{s}^{(p)}_i-\mathbf{s}^{(p)'}_i$ indicates the difference between the original representation and the \textbf{hypothesis}-information enhanced representation of \textbf{premise}, and $\mathbf{s}^{(p)}_i \odot \mathbf{s}^{(p)'}_i$ represents their semantic similarity. Both values are designed to measure the degree of semantic relevance between the sentences pair. The smaller the difference and the larger the semantic similarity, the sentences pair are more likely to be classified into Entailment category. The difference and element-wise product are then concatenated with the original vectors ($\mathbf{S}^{(p)}, \mathbf{S}^{(p)^\prime}$). We expect that such operations could help enhance the pair-level information and capture the inference relationships of \textbf{premise} and \textbf{hypothesis}. We get the new representation containing \textbf{hypothesis}-guided inferential information for \textbf{premise}:
\begin{gather}
  \tilde{\mathbf{S}}^{(p)} = (\tilde{\mathbf{s}}^{(p)}_1, \tilde{\mathbf{s}}^{(p)}_2, \dots, \tilde{\mathbf{s}}^{(p)}_m),\\
  \hat{\mathbf{S}}^{(p)} = LayerNorm(\tilde{\mathbf{S}}^{(p)}),
\end{gather}
where $LayerNorm(.)$ is a layer normalization. The result $\hat{\mathbf{S}}^{(p)}$ is a 2D-tensor that has the same shape as $\mathbf{S}^{(p)}$. The representation of \textbf{hypothesis} $\hat{\mathbf{S}}^{(h)}$ is calculated in the same way. We aggregate these representations and the pair-level representation $\mathbf{Z}$ for the sentence pair is obtained as follows:
\begin{equation}
  \mathbf{Z} = [\hat{\mathbf{S}}^{(p)}; \hat{\mathbf{S}}^{(h)}; \hat{\mathbf{S}}^{(p)} - \hat{\mathbf{S}}^{(h)}; \hat{\mathbf{S}}^{(p)} \odot \hat{\mathbf{S}}^{(h)}],
\end{equation}

As described, the cross attention module can capture the relevance of the sentence pair and well characterize the relationship. Therefore, the pair-level representation can perceive the class information of sentence pairs.

\subsection{Training Objective}

\textbf{Supervised contrastive loss}
A contrastive loss brings the latent representations of samples belonging to the same class closer together, by deﬁning a set of positives (that should be closer) and negatives (that should be further apart). In \cite{khosla2020supervised}, the authors extended the above loss to a supervised contrastive loss by regarding the samples belonging to the same class as positive set. Inspired by this, we adopt supervised contrastive learning objective to align the pair-level representation obtained from cross attention module to distinguish sentence pairs from different classes. 

In the training stage, we randomly sample a batch $\mathcal{I}$ of $K$ examples $(X^{(p)}, X^{(h)}, y)_{i \in \mathcal{I} = \{1, \dots, K\}} $ as denoted in Section \ref{sec:encoder}. We denote the set of positives as $\mathcal{P} = \{p: p\in \mathcal{I}, y_p=y_i\wedge p \neq i\}$, with size $|\mathcal{P} |$. The supervised contrastive loss on the batch $\mathcal{I} $ is defined as:
\begin{align}
  \ell_{i, p} = \frac{\operatorname{exp}(\mathbf{Z}_i \cdot \mathbf{Z}_p/\tau)}{\sum_{k \in \mathcal{I}/i}\operatorname{exp}(\mathbf{Z}_i \cdot \mathbf{Z}_k/\tau)},\\
  \mathcal{L}_{SCL} = \sum_{i \in \mathcal{I}}-\operatorname{log} \frac{1}{|\mathcal{P}|}\sum_{p \in \mathcal{P}} \ell_{i, p},
\end{align}

where $\ell_{i, p}$ indicates the likelihood that pair $i$ is most similar to pair $p$ and $\tau$ is the temperature hyper-parameter. Larger values of $\tau$ scale down the dot-products, creating more difﬁcult comparisons. $\mathbf{Z}_i$ is the pair-level representation of pair $(X^{(p)}, X^{(h)})_{i}$ from the cross attention module. Supervised contrastive loss $\mathcal{L}_{SCL}$ is calculated for every sentence pair among the batch $\mathcal{I}$. To minimize contrastive loss $\mathcal{L}_{SCL} $, the similarity of pairs in the same class should be as large as possible, and the similarity of negative examples should be as small as possible. In this way, we can map positive pairs closer together in the embedding space, while pushing apart negative pairs. 

\textbf{Cross-entropy loss} Supervised contrastive loss mainly focuses on separating each pair apart from the others of different classes, whereas there is no explicit force in discriminating contradiction, neutral and entailment. Therefore, we adopt the softmax-based cross-entropy to form the classfication objective:
\begin{equation}
  \mathcal{L}_{CE} = CrossEntropy(\mathbf{W}\mathbf{Z} + \mathbf{b}, y),
\end{equation}
where $\mathbf{W}$ and $\mathbf{b}$ are trainable parameters. $\mathbf{Z}$ is the pair-level representation from the cross attention module and $y$ is the corresponding label of the pair.

\textbf{Overall loss} The overall loss is a weighted average of CE and the SCL loss, denoted as:
\begin{equation}
  \mathcal{L} = \mathcal{L}_{SCL} + \alpha\mathcal{L}_{CE},
\end{equation}
where $\alpha$ is a hyper-parameter to balance two objectives.

\section{EXPERIMENTAL SETUP}
\label{sec:results}
\subsection{Benchmark Dateset}

We conduct our experiments on NLI task and other 7 transfer learning tasks.

\textbf{Natural language inference task:} We evaluate on two popular benchmarks: the Stanford Natural Language Inference (SNLI) \cite{bowman-etal-2015-large} and the MultiGenre NLI Corpus (MultiNLI) \cite{williams2018broad} and compute classiﬁcation accuracy as the evaluation metric. Detailed statistical information is shown in Table~\ref{table2}.

\begin{table}[ht]
  \centering
\resizebox{0.46\textwidth}{!}{%
  \begin{tabular}{lccccc}
  \hline
  \textbf{Dataset} & \textbf{Train} & \textbf{Dev}  & \textbf{Test} & \textbf{Len(P)} & \textbf{Len(H)} \\ 
  \hline
  SNLI     & 549K  & 9.8K & 9.8K & 14     & 8         \\
  MultiNLI(m) & \multirow{2}{*}{392K}  & 9.8K & 9.8K & 22     & 11        \\
  MultiNLI(mm) &   & 9.8K & 9.8K & 22     & 11   \\ 
  \hline
  \end{tabular}%
  }
  \caption{Statistics of datasets: SNLI, MultiNLI. Len(P) and Len(H) refer to the average length of two sentences respectively. MultiNLI(m) and MultiNLI(mm) indicate the matched and mismatched datasets respectively.}\label{table2}
\end{table}

\textbf{Transfer tasks:} We also evaluate on the following transfer tasks: MR \cite{pang2005lillian}, CR \cite{hu2004mining}, SUBJ \cite{pang2004sentimental}, MPQA \cite{wiebe2005annotating}, SST-2 \cite{socher2013recursive}, TREC \cite{voorhees2000building} and MRPC \cite{dolan2005automatically}. 
For single-sentence classification tasks, we train a logistic regression classifier on top of frozen BERT encoder representation $\mathbf{S}$. In MRPC task, we use the pair-level representation $\mathbf{Z}$ obtained from the cross attention module for the sentence pair to map the semantic space. We follow default conﬁgurations from SentEval \cite{conneau2018senteval}.

\subsection{Implementation Details}
We start from pre-trained checkpoints of BERT \cite{devlin2019bert} (uncased) or RoBERTa \cite{liu2019roberta} (cased). We implement PairSCL based on Huggingface's \texttt{transformers} package~\cite{wolf2019huggingface}. All experiments are conducted on 5 Nvidia GTX 3090 GPUs. 

We train our models for 10 epochs with a batch size of 512 and temperature $\tau$ = 0.05 using an Adam optimizer \cite{kingma2014adam}. The hyper-parameter $\alpha$ is set as 1 for combining objectives. The learning rate is set as 5e-5 for base models and 1e-5 for large models. The maximum sequence length is set to 128.

\subsection{Baseline Models}
To analyze the effectiveness of PairSCL on NLI, we select ESIM~\cite{chen-etal-2017-enhanced}, KIM~\cite{chen2018neural}, ADIN~\cite{liang2019asynchronous}, BERT~\cite{devlin2019bert} and RoBERTa \cite{liu2019roberta} as baselines. They are all trained with NLI supervision.

For transfer tasks, we evaluate with SBERT, SRoBERTa \cite{reimers2019sentence} and SimCSE \cite{gao2021simcse}. We directly report the results from \cite{reimers2019sentence}, since our evaluation setting is the same with theirs.

\begin{table}[h]
  \centering
  \resizebox{0.48\textwidth}{!}{%
  \begin{tabular}{lccc}
  \hline
  \textbf{Model} & \textbf{SNLI} & \textbf{MultiNLI(m)} & \textbf{MultiNLI(mm)} \\ \hline
  ESIM  & 88.0 & 72.3 & 72.1 \\
  KIM  & 88.6 & 77.2 & 76.4 \\
  ADIN  & 88.8  & 78.8 & 77.9 \\ \hline
  BERT   & 89.8 & 83.3 & 82.7 \\
  PairSCL-BERT$_\texttt{base}$ & \textbf{91.9} & \textbf{85.5} & \textbf{84.6} \\ \hline
  RoBERTa$_\texttt{base}$   & $91.2$ & $90.8$ & $90.2$ \\ 
  PairSCL-RoBERTa$_\texttt{base}$ & \textbf{93.2} & \textbf{92.7} & \textbf{92.3} \\ \hline
  \end{tabular}
  }
  \caption{Performance on the test dataset. MultiNLI(m) and MultiNLI(mm) represent the accuracy on matched and mismatched datasets. The best performance is in \textbf{bold} among models with the same pre-trained encoder.}\label{table3}
\end{table}
\vspace{-0.1in}

\begin{table*}[!htbp]
  \centering
  \begin{tabular}{lcccccccc}
  \hline
  \textbf{Model} & \textbf{MR} & \textbf{CR} & \textbf{SUBJ} & \textbf{MPQA} & \textbf{SST} & \textbf{TREC} & \textbf{MRPC} & \textbf{Avg.} \\ \hline
  SBERT$_\texttt{base}$$^\clubsuit $ & 83.64 & 89.43 & 94.39 & 89.86 & 88.96 & 89.60 & 76.00 & 87.41 \\
  SimCSE-BERT$_\texttt{base}$$^\heartsuit $ & 82.69 & 89.25 & 94.81 & 89.59 & 87.31 & 88.40 & 73.51 & 86.51 \\
  PairSCL-BERT$_\texttt{base}$ & \textbf{83.80} & \textbf{89.69} & \textbf{94.94} & \textbf{89.92} & \textbf{89.23} & \textbf{89.87} & \textbf{76.67} & \textbf{87.73} \\ \hline
  SRoBERTa$_\texttt{base}$$^\heartsuit $ & 84.91 & 90.83 & 92.56 & 88.75 & 90.50 & 88.60 & 78.14 & 87.76 \\
  SimCSE-RoBERTa$_\texttt{base}$$^\heartsuit $ & 84.92 & 92.00 & 94.11 & 89.82 & 91.27 & 88.80 & 75.65 & 88.08 \\
  PairSCL-RoBERTa$_\texttt{base}$ & \textbf{85.38} & \textbf{92.67} & \textbf{95.12} & \textbf{90.56} & \textbf{92.14} & \textbf{89.06} & \textbf{79.65} & \textbf{89.23} \\ \hline
  \end{tabular}
  \caption{Transfer task results of different snetence embedding models (measured as accuracy). $\clubsuit$: results from \cite{reimers2019sentence}; $\heartsuit $: results from \cite{gao2021simcse}. The best performance is in \textbf{bold} among models with the same pre-trained encoder.}\label{table4}
\end{table*}
\vspace{-0.1in}

\section{Experiment Results}
\label{sec:experiment}

\subsection{NLI Results}

Table \ref{table3} shows the average results of different models on NLI task. Our approach outperforms state-of-the-art models by 2.1\% on SNLI with BERT encoders. On MultiNLI dataset, we compare on two test sets (matched and mismatched). PairSCL-BERT$_\texttt{base}$ achieves 85.5\% and 84.6\% respectively. For RoBERTa encoders, PairSCL achieves outstanding performance, 93.2\% on SNLI and 92.7\%/92.3\% on MultiNLI. For the results on two datasets, we conduct the students paired t-test and the p-value of the significance test between the results of PairSCL and RoBERTa is less than 0.01 and 0.05, respectively.

This performance gains are due to the stronger ability of PariSCL to learn pair-level representation with cross attention. PairSCL can capture pair-level semantics effectively by the specifically-designed contrastive signal -- predicting whether two sentence pairs belong to the same class.

\subsection{Transfer Tasks Results}

Table \ref{table4} shows the evaluation results on transfer tasks. We can observe that PairSCL-BERT$_\texttt{base}$ outperforms several supervised baselines like InferSent and Universal Sentence Encoder, and keeps comparable to the strong supervised method SBERT$_\texttt{base}$. When further performing representation transfer with RoBERTa base architecture, our approach achieves even better performance. On average, our approach outperforms SimCSE-RoBERTa$_\texttt{base}$ with an improvement of 1.15\% (from 88.08\% to 89.23\%).

As we argued earlier, it benefits from that our model can distinguish the sentences of different classes well by pulling the sentence from the same class together and pushing them of different classes further apart.

\subsection{Ablation Study}
\label{ablation}
To better understand the contribution of each key component of PairSCL, we conduct an ablation study on SNLI based on BERT encoders. The results are shown in Table \ref{table5}. 

After removing the cross attention mechanism, the model simply concat the representation of two sentences. The performance decreases by 1.6\% on the test set which shows the joint representation obtained by cross attention can well characterize the relationship between the sentence pair. 
Remove the cross-entropy loss and the test accuracy decreases by 0.7\%. Without the supervised contrastive learning loss, the accuracy of our model is decreased to 90.7\%. The reason is that the contrastive learning objective can learn the discrepancy between the sentence pairs of different classes by pulling the sentence pairs from the same class together and pushing the pairs of different classes further apart.

\begin{table}[ht]
  \centering
  \begin{tabular}{lc}
  \hline
  \textbf{Model}              & \textbf{Accuracy} \\ \hline
  PairSCL (-CE loss)     & 91.2     \\
  PairSCL (-SCL loss) & 90.7 \\
  PairSCL (-Cross attention) & 90.3     \\
  PairSCL      & \textbf{91.9}   \\ \hline
  \end{tabular}
  \caption{Ablation study on SNLI with BERT encoders.}\label{table5}
\end{table}
\vspace{-0.1in}

\section{CONCLUSION}
\label{sec:conclusions}
In this paper, we propose a pair-level supervised contrastive learning approach. We adopt a cross attention module to learn the joint representations of the sentence pairs. A contrastive learning objective is designed to distinguish the varied classes of sentence pairs by pulling those in one class together and pushing apart the pairs in other classes. We evaluate PairSCL on two popular datasets: SNLI and MultiNLI. The experiment results show that PairSCL obtains new state-of-the-art performance compared with existing models. For the transfer tasks, PariSCL outperforms the previous state-of-the-art method with 1.2\% averaged improvement. We carefully study the components of PairSCL, and show the effects of different parts.

\section{Acknowledgments}
The work was supported by the National Key Research and Development Program of China (No. 2019YFB1704003), the National Nature Science Foundation of China (No. 62021002 and No. 71690231), Tsinghua BNRist and Beijing Key Laboratory of Industrial Big Data System and Application.
\vfill
\pagebreak

\ninept 

\bibliographystyle{IEEEbib}
\bibliography{refs}

\end{document}